\documentclass[letterpaper]{article} 
\usepackage{aaai2027}  

\usepackage[hyphens]{url}  
\usepackage{graphicx} 
\usepackage{natbib}  
\usepackage{caption} 
\usepackage{algorithm}
\usepackage{multirow}
\usepackage{algorithmic}
\usepackage{newfloat}
\usepackage{listings}
\usepackage{booktabs}
\usepackage{multirow}
\usepackage{graphicx}
\usepackage{xcolor}

\usepackage{colortbl}

\DeclareCaptionStyle{ruled}{labelfont=normalfont,labelsep=colon,strut=off} 

\usepackage{booktabs}
\usepackage{tabularx}
\usepackage{array}
\usepackage{makecell}
\usepackage{multirow}
\usepackage{pifont}
\usepackage{xcolor}

\definecolor{checkgreen}{RGB}{0,0,0}
\definecolor{crossred}{RGB}{0,0,0}

\definecolor{resultgreen}{RGB}{0,130,0}
\definecolor{resultred}{RGB}{200,0,0}

\definecolor{preferrorblue}{RGB}{255, 255, 255}
\definecolor{taskerrorgreen}{RGB}{255, 255, 255}
\definecolor{grayfill}{RGB}{255, 255, 255} 

\definecolor{improvedfill}{RGB}{255, 255, 255}
\definecolor{notimprovedfill}{RGB}{255, 255, 255}

\definecolor{graytext}{RGB}{90,90,90} 

\newcommand{\preferrorcell}[1]{\cellcolor{preferrorblue}#1}
\newcommand{\taskerrorcell}[1]{\cellcolor{taskerrorgreen}#1}
\newcommand{\jointerrorcell}[1]{\cellcolor{grayfill}#1}

\newcommand{\cmark}{\Large\textcolor{checkgreen}{\ding{51}}}
\newcommand{\xmark}{\Large\textcolor{crossred}{\ding{55}}}

\newcommand{\brow}[5]{#1 & #2 & #3 & #4 & #5 \\}

\floatstyle{ruled}
\newfloat{listing}{tb}{lst}{}
\floatname{listing}{Listing}
\usepackage{booktabs}
\title{PDEU-Bench: Benchmarking the Personalized Planning Lifecycle of Tool-Calling LLM Agents}
\author{
Huayi Lai\textsuperscript{1},
Shichao Song\textsuperscript{1},
Qingchen Yu\textsuperscript{2},
Simin Niu\textsuperscript{1},
Mengwei Wang\textsuperscript{1},
Hanyu Wang\textsuperscript{1},
Xun Liang\textsuperscript{1}\corresponding
}
\affiliations{
\textsuperscript{\rm 1}Renmin University of China\\
\textsuperscript{\rm 2}Beihang University\\
xliang@ruc.edu.cn
}
\begin{document}
\maketitle

\begin{abstract}
Large language model (LLM) agents are evolving from tool-calling systems that execute isolated instructions into task-oriented agents that pursue user goals through sustained, multi-step interactions. However, existing benchmarks for personalized tool use largely assess isolated calls or reactive execution, leaving unclear whether agents can formulate, execute, and revise an explicit plan while preserving user preferences throughout long-term interaction. To address this gap, we introduce \textbf{PDEU-Bench} (\textbf{P}ersonalized plan \textbf{D}efinition, plan \textbf{E}xecution, and plan \textbf{U}pdate \textbf{Bench}mark), a benchmark for evaluating the complete planning lifecycle of personalized tool-using agents. PDEU-Bench comprises 214 long-horizon interaction tasks spanning 12 everyday domains and 94 tools, with stage-specific assessments of preference adherence and plan quality. Extensive evaluations of 15 representative open-source and closed-source LLMs reveal a pronounced gap between local tool execution and dynamic planning: LLMs can often instantiate preferences in individual calls, yet struggle to construct coherent plan definition and plan update. We further evaluate mainstream personalization and memory-augmentation methods. Although these methods improve particular stages, none of the evaluated methods reliably propagates user preferences throughout the complete lifecycle, and their gains frequently fail to transfer to subsequent execution. Fine-grained error analysis further reveals that preference omissions and conflicts persist throughout the planning lifecycle, highlighting the need for future research to parameterize LLMs with preference-aware information retrieval and memory capabilities. We provide the relevant code and data in the appendix to support future research.
\end{abstract}

\section{Introduction}
Personalized agentic tool use has become a foundational capability of large language models (LLMs). Traditional personalization primarily leverages user information to generate responses~\cite{prefeval} or recommendations aligned with individual preferences~~\cite{agentrecbench,recbench}. In contrast, emerging LLM agents are expected to make personalized decisions~~\cite{ptbench,ToolSpectrum}, formulate plans~\cite{deepplanning}, and execute actions within specific interactive environments~\cite{planbenchxl}. Agents must interpret high-level, underspecified requests, infer latent user preferences from interaction histories, and continually revise their action plans as tool outputs reshape the space of feasible next steps. 

Successful task completion therefore requires more than selecting appropriate tools and producing syntactically valid arguments. Agents must decompose complex objectives, coordinate subgoals and tool dependencies, translate personalized constraints into executable actions, and adapt their plans to environmental feedback\cite{vitabench2}. Without effective planning, long tool-use trajectories are vulnerable to cascading errors~\cite{planbenchxl}, redundant interactions that inflate cost and latency~\cite{LessContext,WhentoCall}, and gradual drift from user-specific constraints and the intended goal~\cite{ShoppingCompanion,compass}.

\begin{figure}[t]
    \centering
    \includegraphics[width=1\linewidth]{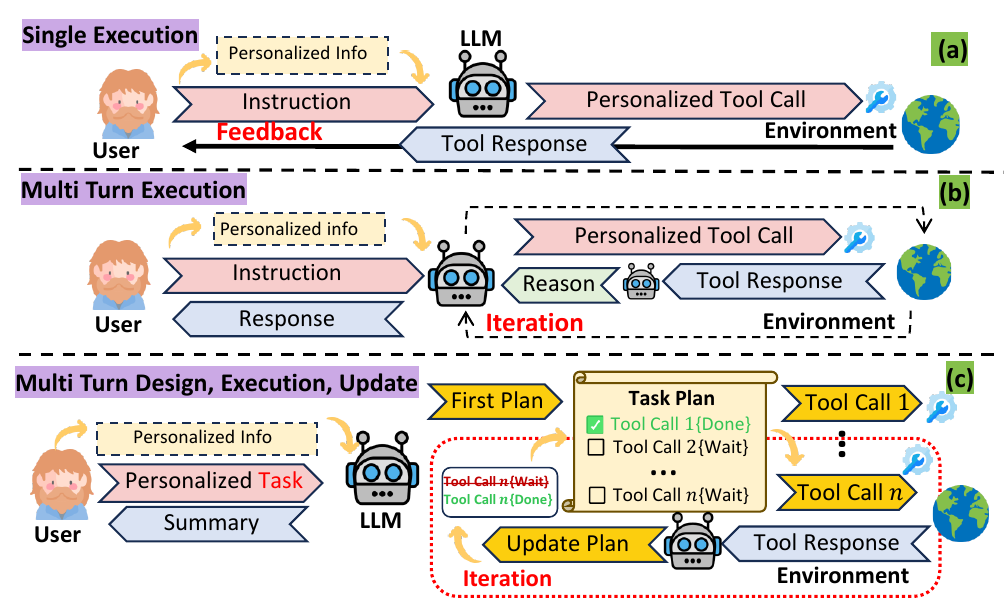}
    \caption{Motivation for PDEU-Bench: a benchmark for personalized tool-use planning across the whole planning lifecycle.
    (a) Single-turn evaluation focuses on an isolated personalized single execution.
    (b) Multi-turn reactive evaluation extends interaction through repeated reasoning and plan execution.
    (c) PDEU-Bench treats the plan as an explicit, evolving object and evaluates its core lifecycle through plan definition, plan execution (Tool Call), and feedback-driven plan update.}
    
    \label{fig:motivation}
\end{figure}
As illustrated in part(a) and part(b) in Figure~\ref{fig:motivation}, existing personalized tool-use benchmarks largely focus on either single-turn execution, which evaluates an isolated personalized tool call~\cite{AstraBench,ToolSpectrum}, or multi-turn reactive execution, which extends interaction through repeated reasoning and tool execution~\cite{userbench,FamilyTool}. Although the latter increases the interaction horizon, both settings focus solely on task execution; neither treats an explicit and evolving plan as the primary evaluation object. This raises a central question: \textbf{Can an agent formulate, execute, and continually update an actionable plan while remaining executable and aligned with user preferences?}

To address this gap, we design and develop \textbf{PDEU-Bench} (\textbf{P}ersonalized plan \textbf{D}efinition, plan \textbf{E}xecution, \textbf{U}pdate \textbf{Bench}mark), a multi-stage agentic benchmark for evaluating the planning lifecycle in personalized tool-use tasks, which \textbf{systematically evaluates LLM's personalized plan definition, plan execution, and plan update capabilities in multi-round tool-calling interactions}. As shown in Figure~\ref{fig:motivation}(c), PDEU-Bench treats the plan as an explicit and evolving evaluation object and decomposes its lifecycle into three interconnected yet independently measurable stages:
\begin{itemize}
    \item \textbf{Plan Definition:} which evaluates whether an agent can combine an underspecified user request with personalized context to formulate a structured, executable, and preference-aligned initial plan;
    \item \textbf{Plan Execution:} which evaluates whether an agent can select appropriate tools according to the plan and accurately instantiate task constraints and user preferences as tool-call arguments;
    \item \textbf{Plan Update:} which evaluates whether an agent can revise affected subsequent steps in response to tool outputs while preserving goals, dependencies, and preferences that remain valid.
\end{itemize}

PDEU-Bench comprises 214 personalized scenarios across 12 everyday domains and 94 tools, integrating user goals, personalized contexts, tool interfaces, and executable environments into unified task trajectories. By separately evaluating plan definition, plan execution, and plan update, it extends personalized tool-use evaluation from isolated calls to the complete planning lifecycle. Using PDEU-Bench, we evaluate 15 mainstream open-source and closed-source LLMs, revealing performance deficiencies in mainstream LLMs and highlighting a clear gap between planning capabilities and tool execution capabilities. Furthermore, we evaluated six preference-enhancement methods and found that existing approaches exhibit significant performance limitations. Our main contributions are summarized as follows:
\begin{itemize}
    \item We introduce \textbf{PDEU-Bench}, an open-source benchmark for \textbf{personalized tool-use planning across plan definition, plan execution, and plan update}, covering 214 scenarios, 12 domains, and 94 tools.

    \item We evaluate 15 representative open-source and closed-source LLMs, revealing a clear gap between plan execution and lifecycle-level planning, with personalized plan definition and feedback-driven plan updating are key bottlenecks.

    \item We systematically evaluate six personalization and memory augmentation strategies and find that none improves performance across all three stages simultaneously, revealing a general limitation in preserving user preferences throughout the planning lifecycle.
\end{itemize}

\section{Related Work}
\begin{table}[t]
\centering
\small
\setlength{\tabcolsep}{1.2pt}
\renewcommand{\arraystretch}{1.05}

\begin{tabularx}{\columnwidth}{
  @{}
  >{\centering\arraybackslash}m{0.26\columnwidth}
  @{\hspace{1pt}}
  >{\centering\arraybackslash}m{0.18\columnwidth}
  @{\hspace{2pt}\vrule width \arrayrulewidth\hspace{2pt}}
  *{3}{>{\centering\arraybackslash}X}
  @{}
}
\toprule

\multirow{2}{*}{
  \makecell[c]{
    \textbf{Existing}\\
    \textbf{Benchmarks}
  }
}
&
\multirow{2}{*}{
  \makecell[c]{
    \textbf{Multi-turn}\\
    \textbf{Planning}
  }
}
&
\multicolumn{3}{c}{\textbf{Plan Action}}
\\

\cmidrule(lr){3-5}

& 
& \textbf{Definition}
& \textbf{Execution}
& \textbf{Update}
\\

\midrule

\brow{PTBench}
     {\xmark}{\xmark}{\cmark}{\xmark}

\brow{ToolSpectrum}
     {\xmark}{\xmark}{\cmark}{\xmark}

\brow{PEToolBench}
     {\xmark}{\xmark}{\cmark}{\xmark}

\brow{Claw-Anything}
     {\xmark}{\xmark}{\cmark}{\xmark}

\brow{ASTRA-Bench}
     {\xmark}{\xmark}{\cmark}{\xmark}

\brow{MPT}
     {\xmark}{\xmark}{\cmark}{\xmark}

\brow{DeepPlanning}
     {\xmark}{\cmark}{\xmark}{\xmark}

\brow{TravelBench}
     {\xmark}{\cmark}{\xmark}{\xmark}

\midrule

\brow{Family Tool}
     {\cmark}{\xmark}{\cmark}{\xmark}

\brow{UserBench}
     {\cmark}{\xmark}{\cmark}{\xmark}

\brow{ETAPP}
     {\cmark}{\xmark}{\cmark}{\xmark}

\brow{GroupTravelBench}
     {\cmark}{\xmark}{\cmark}{\xmark}


\brow{MCP-Persona}
     {\cmark}{\xmark}{\cmark}{\xmark}

\brow{PersonalWAB}
     {\cmark}{\xmark}{\cmark}{\xmark}

\brow{APOLLO}
     {\cmark}{\xmark}{\cmark}{\xmark}

\brow{VitaBench2.0}
     {\cmark}{\xmark}{\cmark}{\xmark}

\midrule

\brow{\textbf{PDEU-Bench}}
     {\cmark}{\cmark}{\cmark}{\cmark}

\bottomrule
\end{tabularx}
\caption{Comparison of existing benchmarks in terms of multi-turn planning and plan action capabilities.}
\label{tab:related_works}
\end{table}
\begin{figure*}[t]
    \centering
    \includegraphics[width=\textwidth]{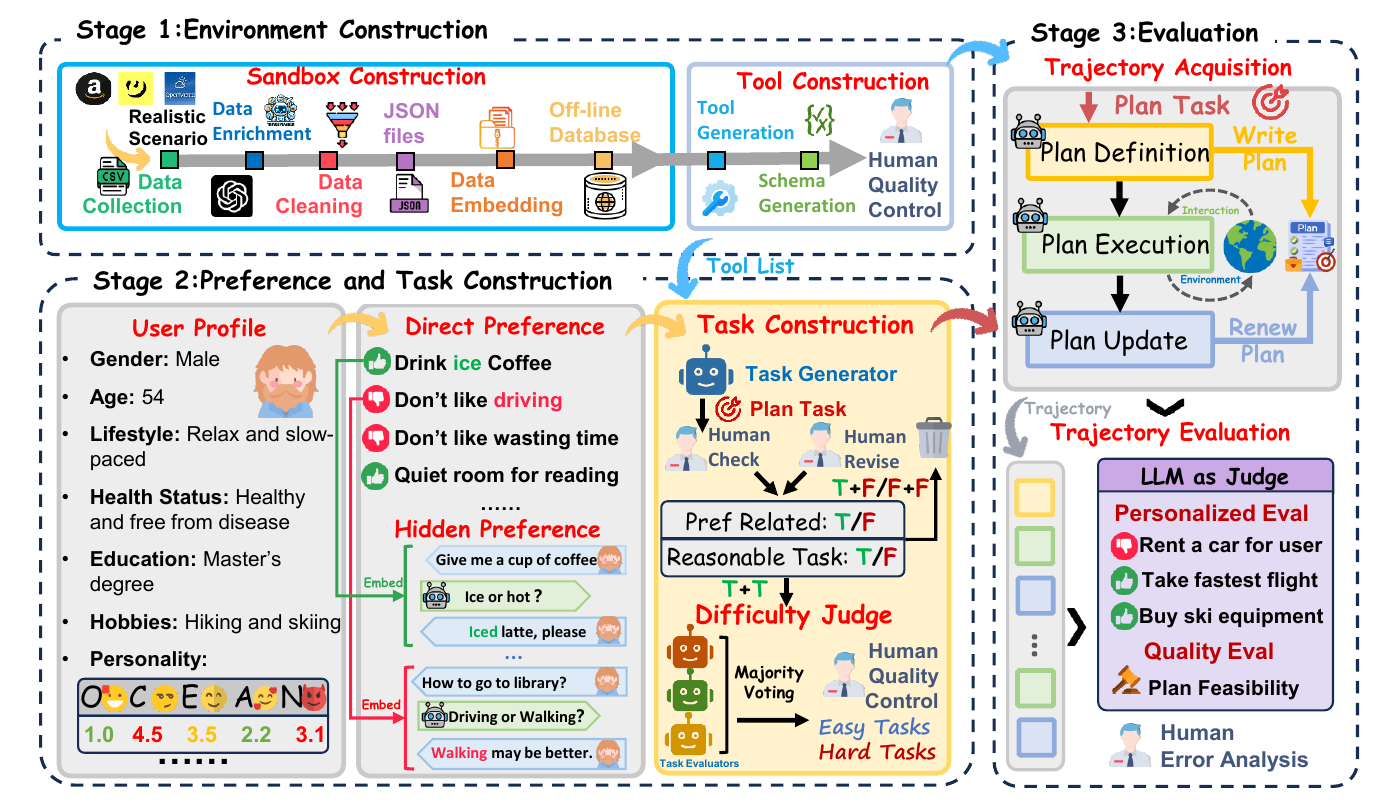}
    \caption{Overview of the proposed benchmark PDEU-Bench}
    \label{fig:framework}
\end{figure*}

\subsection{Benchmarks for LLM Personalized Tool Use}

As shown in Table~\ref{tab:related_works}, existing benchmarks for personalized tool-using LLMs largely focus on either task execution or isolated planning capabilities. Single-turn benchmarks such as PTBench\cite{ptbench}, ToolSpectrum\cite{ToolSpectrum}, PEToolBench\cite{PETooBench}, Claw-Anything\cite{clawanything}, ASTRA-Bench\cite{AstraBench}, and MPT\cite{MPT} mainly evaluate whether agents can infer user preferences and issue valid tool calls, while DeepPlanning\cite{deepplanning} assesses plan generation without execution or revision. Multi-turn benchmarks, including FamilyTool\cite{FamilyTool}, UserBench\cite{userbench}, ETAPP\cite{etapp}, GroupTravelBench\cite{GroupTravelBench}, MCP-Persona\cite{MCP-Persona}, PersonalWAB\cite{personalwab}, APOLLO\cite{apollo}, and VitaBench2.0\cite{vitabench2}, extend evaluation to longer interaction trajectories but still center on task completion rather than plan definition. TravelBench\cite{TravelBench} and DeepPlanning\cite{deepplanning} evaluated plan definition capabilities but were limited to single-round plan setting. In contrast, PDEU-Bench is an agentic benchmark to evaluate personalized tool-using LLMs throughout the entire planning lifecycle, jointly covering plan definition, execution, and feedback-driven update while measuring whether user preferences and plan quality are maintained consistently across stages.

\subsection{Benchmarks for LLM Personalized Planning}

LLM-based personalized planning can be divided into \textit{role-playing planning} and \textit{preference-following planning}. \textit{Role-playing planning} primarily focuses on evaluating whether the decisions and plans made by role-playing LLM regarding role-related tasks align with the role's requirements\cite{rolecde,ceobench}. \textit{Preference-following planning} instead aims to align plans and actions with users' explicit or implicit preferences. However, existing work in this direction exhibits strong domain specificity, focusing on settings such as e-commerce shopping \cite{ShopperBench,deepplanning}, travel \cite{travelplanner,triptailor,travelplanner++,Triptide}, healthcare\cite{medplan},household\cite{household} or sports\cite{planfitting}. These benchmarks lack a general purpose evaluation across diverse user needs and typically assess plan definition or plan execution in isolation, without systematically evaluating plan update. In contrast, PDEU-Bench comprises 214 personalized planning tasks spanning 12 everyday domains, it is a comprehensive multi-domain agentic benchmark for evaluating the full planning lifecycle in personalized planning, covering plan definition, tool-grounded plan execution, and feedback-driven plan update.

\section{PDEU Benchmark}

As shown in Figure \ref{fig:framework}, PDEU-Bench includes 3 components: Environment Construction(Stage 1), Preference , Task Construction(Stage 2) and Trajectory Evaluation(Stage 3).
\subsection{Environment Construction}
\paragraph{Sandbox Construction} Considering the network fluctuations and high maintenance costs associated with online evaluation environments, we develop an extensible offline sandbox to ensure the stability of the evaluation process and the reproducibility of the results. As shown in Figure~\ref{fig:framework} , the sandbox isolates experiments from external environmental variables, preventing external factors from affecting API outputs and thereby ensuring evaluation accuracy.  Real-world API data is used to generate corresponding outputs (e.g., weather from Open-Meteo Forecast API), enhancing the fidelity of the simulated scenarios. To further enrich the number of available tools and simulate real-world complexities, gpt-5.5 is used to generate data on news topics. Statistical analysis showed that 85.7\% of the data was based on realistic scenarios. For further details, please refer to appendix.

After data enrichment, we deterministically clean and normalize the heterogeneous records, and serialize them into a unified JSONL schema with provenance metadata and structured attributes. We then pre-encode the records using "multilingual-e5-small"~\cite{e5model} to construct an offline semantic index. The resulting sandbox integrates external evidence with deterministic execution states, enabling reproducible tool observations and feedback throughout evaluation.
\begin{figure}
    \centering
    \includegraphics[width=1\linewidth]{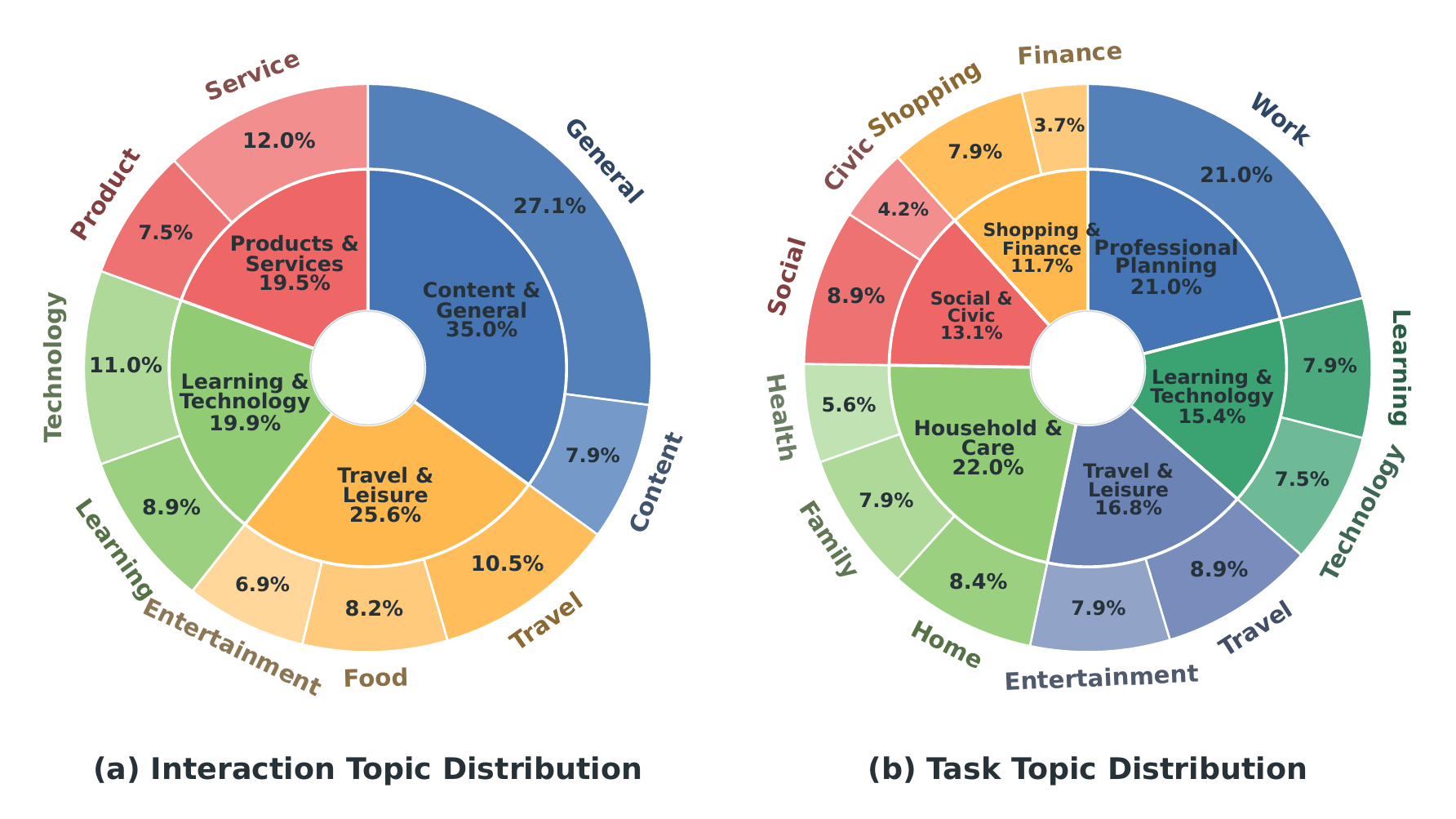}
    \caption{Six high-level planning
categories and the proportion of individual task types within each category.}
    \label{fig:topic}
\end{figure}
\paragraph{Tool Construction}
To support realistic interaction with the offline sandbox, we construct 94 functional APIs supporting 12 everyday domains and covering both information-seeking and state-changing operations. Following the OpenAI function-calling specification\cite{schemasurvey}. GPT-5.5\cite{gpt5} is used to generate each API schema with a standardized structure and preference-sensitive parameters designed to instantiate task-relevant user preferences. The generated schemas are then assigned to two professional annotators for manual verification of functional correctness and the inclusion of preference-sensitive parameters. Schemas that fail either criterion are revised and rechecked. 

\subsection{Preference and Task Construction} Long-term interaction histories typically reflect stable user profiles and traits at a coarse level, while personalized preferences are implicitly expressed in individual interactions~\cite{vitabench2}. As illustrated in Figure~\ref{fig:framework}, a multi-stage automated generation pipeline was designed and developed for simulation.

\paragraph{Preference Construction}Preference data were constructed to capture context-dependent user preferences implicitly expressed across interaction histories. First, 300 persona seeds were sampled from PersonaHub~\cite{personahub} and expanded into coherent fictional profiles (e.g., big five personality traits and hobbies). Guided by preference construction theory~\cite{prefconstruction}, multidimensional preferences were instantiated across ten everyday domains (e.g., like of iced coffee in food domain). Following PrefEval~\cite{prefeval}, these preferences were implicitly conveyed through 15-25 two-turn episodic interactions per user using experiences, choices, and consequences. Consistency among profiles, preferences, and interactions was reviewed by three professional annotators, with invalid instances revised or discarded. Overall, 83.0\% of instances were retained unchanged and 17.0\% after minor revision; none were discarded.

\paragraph{Task Construction} For each user, a multi-step planning task was generated based on the corresponding profile, direct preferences, and available tool set. Preference descriptions were not explicitly exposed in the user request. Each candidate was then manually assessed by 3 professional annotators using two binary criteria: \emph{preference relevance}, which determined whether task completion required profile-grounded preferences, and \emph{task reasonableness}, which assessed whether the request was coherent and executable with the available tools. Candidate tasks are discarded only when they simultaneously lack sufficient preference relevance and cannot be completed with the available tools (F+F); otherwise, they undergo manual revision and re-evaluation(T+F), in this stage, 45.0\% of the tasks were retained without revision, 41.0
\% were retained after revision, and 14.0\% were discarded.To evaluate the impact of task difficulty on model performance, task difficulty was independently assessed by three heterogeneous LLM evaluators drawn from different model families and configurations: mimo-v2.5-pro, gemini-3.1-Flash, and gpt-5.5. A professional annotator checked the results after majority voting, check appendix for details. Of the retained 214 tasks, 68 were rated as easy and 146 as hard.

\begin{table*}[t!]
\centering
\small
\setlength{\tabcolsep}{4pt}
\renewcommand{\arraystretch}{1.08}
\resizebox{\textwidth}{!}{%
\begin{tabular}{@{}l*{9}{c}@{}}
\toprule
\multirow{2}{*}{\textbf{Models}} & \multirow{2}{*}{\textbf{Pref. Score}} & \multirow{2}{*}{\textbf{Qual. Score}} & \multicolumn{3}{c}{\textbf{Plan Definition}} & \multicolumn{3}{c}{\textbf{Plan Update}} & \textbf{Plan Execution} \\
\cmidrule(lr){4-6}\cmidrule(lr){7-9}\cmidrule(l){10-10}
& & & \textbf{Pref. Acc.} & \textbf{Qual. Acc.} & \textbf{Joint Acc.} & \textbf{Pref. Acc.} & \textbf{Qual. Acc.} & \textbf{Joint Acc.} & \textbf{Pref. Acc.} \\
\midrule
\multicolumn{10}{c}{\textit{Closed-Source Models}} \\
\midrule
gpt-5.6-terra         & \preferrorcell{\textbf{0.6632}} & \taskerrorcell{\textbf{0.4350}} & 0.7908 & \taskerrorcell{\textbf{0.7329}} & \jointerrorcell{\textbf{0.6027}} & \preferrorcell{\textbf{0.3425}} & \taskerrorcell{\textbf{0.1370}} & \jointerrorcell{\textbf{0.1370}} & 0.8562 \\
claude-sonnet-5       & \underline{0.6134} & 0.3219 & 0.7055 & 0.5616 & 0.4452 & \underline{0.2192} & \underline{0.0822} & \underline{0.0753} & 0.9157 \\
grok-4.5              & 0.6112 & \underline{0.3767} & \preferrorcell{\textbf{0.8151}} & \underline{0.7123} & 0.5959 & 0.0858 & 0.0411 & 0.0068 & \preferrorcell{\underline{0.9328}} \\
gpt-5.4               & 0.5551 & 0.3630 & \underline{0.8082} & 0.6918 & \underline{0.5959} & 0.1781 & 0.0342 & 0.0205 & 0.6791 \\
grok-4.3              & 0.5296 & \underline{0.3767} & 0.7466 & \underline{0.7123} & 0.5686 & 0.1164 & 0.0411 & 0.0274 & 0.7260 \\
gpt-5.4-mini          & 0.4680 & 0.2945 & 0.4863 & 0.5205 & 0.2808 & 0.1712 & 0.0685 & 0.0411 & 0.7466 \\
gemini-3.5-flash      & 0.4438 & 0.1849 & 0.2671 & 0.3356 & 0.1712 & 0.1301 & 0.0342 & 0.0274 & \textbf{0.9344} \\
gemini-3.1-flash-lite & 0.4414 & 0.1164 & 0.2329 & 0.1507 & 0.0411 & 0.1849 & 0.0822 & 0.0616 & 0.9066 \\
\midrule
\textbf{Average Performance}         & 0.5408 & 0.3086 & 0.6066 & 0.5522 & 0.4127 & 0.1785 & 0.0651 & 0.0496 & 0.8372 \\
\midrule
\multicolumn{10}{c}{\textit{Open-Source Models}} \\
\midrule
mimo-v2.5-pro         & \preferrorcell{\underline{0.4518}} & 0.2547 & 0.3271 & 0.3551 & 0.1308 & \underline{0.3087} & 0.1542 & 0.1121 & \preferrorcell{0.7196} \\
mimo-v2.5             & 0.4276 & \underline{0.3117} & 0.3904 & 0.2808 & 0.1575 & \preferrorcell{\textbf{0.3384}} & \taskerrorcell{\textbf{0.3425}} & \jointerrorcell{\underline{0.1945}} & 0.5541 \\
deepseek-v4-pro       & 0.4361 & 0.2781 & \preferrorcell{\textbf{0.5421}} & \taskerrorcell{\textbf{0.5000}} & \jointerrorcell{\textbf{0.2991}} & 0.1869 & 0.0561 & 0.0374 & 0.5794 \\
deepseek-v4-flash       & \textbf{0.4714} & 0.2089 & 0.4178 & 0.3699 & 0.1781 & 0.0616 & 0.0479 & 0.0205 & \textbf{0.9347} \\

deepseek-v3.2         & 0.4145 & 0.2363 & \underline{0.4452} & 0.4178 & 0.2192 & 0.0822 & 0.0548 & 0.0411 & 0.7162 \\
deepseek-v3.1         & 0.4328 & \taskerrorcell{\textbf{0.3357}} & 0.3336 & 0.3973 & 0.2397 & 0.2425 & \underline{0.2740} & \textbf{0.2166} & \underline{0.7225} \\
qwen3.5-397b-a17b          & 0.4310 & 0.2671 & 0.4178 & \underline{0.4726} & \underline{0.2534} & 0.1712 & 0.0616 & 0.0342 & 0.7040 \\
\midrule
\textbf{Average Performance}         & 0.4379 & 0.2704 & 0.4106 & 0.3991 & 0.2111 & 0.1988 & 0.1416 & 0.0981 & 0.7044 \\
\bottomrule
\end{tabular}%
}
\caption{Average accuracy across models for plan definition, update, and execution on “easy” and “hard” tasks. \textbf{Bold} and \underline{underlined} values mark the best and second best results.}
\label{tab:plan_performance}
\end{table*}

\subsection{Task Evaluation}

\paragraph{Trajectory Acquisition} As shown in Figure~\ref{fig:framework}, each task is executed under a standardized three-stage protocol in which the plan is maintained as an explicit, versioned state. A schema-validated initial plan is produced before tool use, after which an action--observation--update cycle is enforced; every execution attempt, including failures, must be followed by a plan update. Plan versions, tool calls, observations, and termination states are logged and automatically validated for temporal and structural consistency. Task inputs, tool contracts, and sandbox states are held fixed, enabling reproducible and leakage-free assessment of plan definition, plan execution, and plan update.

\paragraph{Trajectory Evaluation} Each trajectory is evaluated using fine-grained binary rubrics. Plan definition and plan update are assessed for preference adherence and plan quality, with each update judged as a transition conditioned on the preceding plan, action, and observation. Plan execution evaluates only preference-aware tool argument instantiation, while response quality is excluded to avoid conflating agent behavior with environmental availability. Evidence-grounded labels are assigned by a fixed judge and deterministically aggregated into stage-level metrics, with success requiring all applicable criteria to be satisfied without prescribing a reference trajectory. To operationalize lifecycle-level reliability in long-horizon settings, we adopt a strict trajectory-level success criterion, under which an instance is considered successful only if all applicable preference and planning requirements are satisfied at every relevant stage and transition. Failure modes are further analyzed by 4 professional annotators from the complementary perspectives of preference adherence and plan quality.

\paragraph{Evaluation Matrix} Preference adherence is evaluated at \textit{plan definition} (\(D\)), \textit{plan execution} (\(E\)), and \textit{plan update} (\(U\)) using fine-grained binary rubrics. Let \(N\) denote the number of evaluated instances and \(P_i^s\in\{0,1\}\) the preference-pass indicator for instance \(i\) at stage \(s\). For \emph{Plan Definition}, \(P_i^D=1\) only if the initial plan both covers all relevant preferences and avoids preference violations. For \emph{Plan Execution}, \(P_i^E=1\) only if every tool call correctly instantiates the applicable preferences in its arguments. For \emph{Plan Update}, \(P_i^U=1\) only if every update retains valid preferences, replans accordingly, grounds them in subsequent actions, and introduces no violation. Preference Accuracy is computed separately for each stage:
\begin{equation}
    \mathrm{Pref Acc.}_s=\frac{1}{N}\sum_{i=1}^{N}P_i^s,
    \qquad s\in\{D,E,U\}.
\end{equation}
Preference accuracy is displayed as "Pref. Acc." in Table~\ref{tab:plan_performance} and Table~\ref{tab:improvement_methods}. The overall Preference Score is defined as the mean of the three stage-level accuracies.
Plan quality is evaluated at \textit{Plan Definition} and \textit{Plan Update}. Let \(Q_i^s\in\{0,1\}\) denote whether instance \(i\) satisfies all quality criteria at stage \(s\in\{D,U\}\). For \emph{Plan Definition}, six criteria are applied: goal and success-criterion coverage, step actionability, dependency and ordering correctness, tool and information acquisition, constraint feasibility, and global coherence. For \emph{Plan Update}, the corresponding criteria assess remaining-goal coverage, replanning correctness, remaining-step actionability, information acquisition and verification, post-update feasibility, and global coherence. Thus, \(Q_i^D=1\) only if all six initial-plan criteria are passed, whereas \(Q_i^U=1\) only if all six criteria are passed for every update. Plan Quality Accuracy(Qual. Acc.) is defined as
\begin{equation}
    \mathrm{Quality Acc.}_s=\frac{1}{N}\sum_{i=1}^{N}Q_i^s,
    \qquad s\in\{D,U\}.
\end{equation}
The overall Quality Score is the mean of \(\mathrm{QAcc}_D\) and \(\mathrm{QAcc}_U\). No quality score is assigned to \textit{Plan Execution} because tool-return quality can't reflect the correctness of the agent's call. Joint Accuracy(Joint Acc.) is used to calculate the percentage of tasks that pass both preference compliance checks and plan rationality checks. Check appendix for details.

\subsection{Quality Control}
Majority voting based on different parameters and family was adopted to minimize sampling variance and stabilize consensus labels~\cite{temperature1}. As shown in Figure~\ref{fig:topic}, the resulting benchmark contains 214 tasks across six categories, with source interactions spanning 9 user-interest themes. To reduce subjective differences in manual annotation and manual statistics, all human-involved stages, including schema verification, task review and revision, and post-vote quality control and adjudication, followed stage-specific instructions and predefined criteria. Check appendix for more details.
\begin{figure}
    \centering
    \includegraphics[width=1\linewidth]{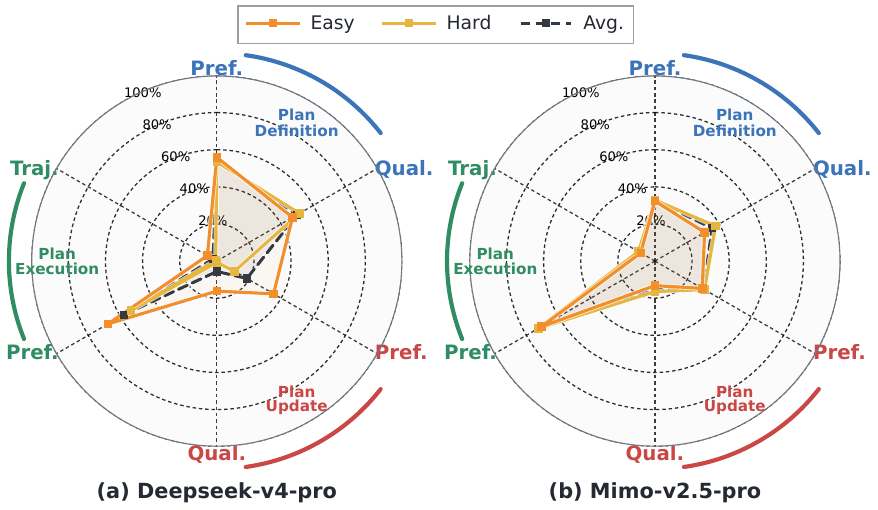}
    \caption{LLM's performance across tasks of varying difficulty. Pref., Qual., and Traj. represent preference adherence accuracy, plan quality accuracy, and task trajectory accuracy, respectively.}
    \label{fig:difficulty}
\end{figure}

\begin{table*}[t]
\centering
\setlength{\tabcolsep}{3.2pt}
\renewcommand{\arraystretch}{1.08}

\resizebox{\textwidth}{!}{%
\begin{tabular}{@{}ll*{9}{c}@{}}
\toprule

\multirow{2}{*}{\textbf{Models}}
& \multirow{2}{*}{\textbf{Methods}}
& \multirow{2}{*}{\textbf{Pref. Score}}
& \multirow{2}{*}{\textbf{Qual. Score}}
& \multicolumn{3}{c}{\textbf{Plan Definition}}
& \multicolumn{3}{c}{\textbf{Plan Update}}
& \textbf{Plan Execution} \\

\cmidrule(lr){5-7}
\cmidrule(lr){8-10}
\cmidrule(lr){11-11}

& & & &
\textbf{Pref. Acc.}
& \textbf{Qual. Acc.}
& \textbf{Joint Acc.}
& \textbf{Pref. Acc.}
& \textbf{Qual. Acc.}
& \textbf{Joint Acc.}
& \textbf{Pref. Acc.} \\

\midrule

\multirow{9}{*}{Gemini}
& Baseline
& 0.4414 & 0.1164
& 0.2329 & 0.1507 & 0.0411
& 0.1849 & 0.0822 & 0.0616
& 0.9066 \\

\cmidrule(lr){2-11}

& Reminder
& 0.3949(\textbf{↓}) & 0.1506(\textbf{↑})
& 0.3219(\textbf{↑}) & 0.2260(\textbf{↑}) & 0.0753(\textbf{↑})
& 0.1644(\textbf{↓}) & 0.0753(\textbf{↓})
& 0.0685(\textbf{↑}) & 0.6986(\textbf{↓}) \\

\cmidrule(lr){2-11}

& \multicolumn{10}{c}{\textit{\textbf{RAG-based Methods}}} \\

& RAG
& 0.4497(\textbf{↑}) & 0.1301(\textbf{↑})
& 0.3836(\textbf{↑}) & 0.2192(\textbf{↑}) & 0.1370(\textbf{↑})
& 0.2123(\textbf{↑}) & 0.0411(\textbf{↓})
& 0.0342(\textbf{↓}) & 0.7534(\textbf{↓}) \\

& PAG
& 0.4908(\textbf{↑}) & 0.2226(\textbf{↑})
& 0.5753(\textbf{↑}) & 0.3699(\textbf{↑}) & 0.2945(\textbf{↑})
& 0.2055(\textbf{↑}) & 0.0753(\textbf{↓})
& 0.0753(\textbf{↑}) & 0.6918(\textbf{↓}) \\

\cmidrule(lr){2-11}

& \multicolumn{10}{c}{\textit{\textbf{Memory-based Methods}}} \\

& MemoryBank
& 0.5414(\textbf{↑}) & 0.2603(\textbf{↑})
& 0.5411(\textbf{↑}) & 0.3562(\textbf{↑}) & 0.2603(\textbf{↑})
& 0.3699(\textbf{↑}) & 0.1644(\textbf{↑})
& 0.1438(\textbf{↑}) & 0.7132(\textbf{↓}) \\

& PersonaAgent
& 0.6118(\textbf{↑}) & 0.3664(\textbf{↑})
& 0.5342(\textbf{↑}) & 0.4863(\textbf{↑}) & 0.4272(\textbf{↑})
& 0.4726(\textbf{↑}) & 0.2466(\textbf{↑})
& 0.2055(\textbf{↑}) & 0.8288(\textbf{↓}) \\

& Agentic Memory
& 0.5570(\textbf{↑}) & 0.4041(\textbf{↑})
& 0.4452(\textbf{↑}) & 0.5411(\textbf{↑}) & 0.3904(\textbf{↑})
& 0.5205(\textbf{↑}) & 0.2671(\textbf{↑})
& 0.2192(\textbf{↑}) & 0.7055(\textbf{↓}) \\

\midrule

\multirow{9}{*}{Deepseek}
& Baseline
& 0.4714 & 0.2089
& 0.4178 & 0.3699 & 0.1781
& 0.0616 & 0.0479 & 0.0205
& 0.9347 \\

\cmidrule(lr){2-11}

& Reminder
& 0.4634(\textbf{↓}) & 0.3082(\textbf{↑})
& 0.4315(\textbf{↑}) & 0.4452(\textbf{↑}) & 0.3014(\textbf{↑})
& 0.3288(\textbf{↑}) & 0.1712(\textbf{↑})
& 0.1301(\textbf{↑}) & 0.6301(\textbf{↓}) \\

\cmidrule(lr){2-11}

& \multicolumn{10}{c}{\textit{\textbf{RAG-based Methods}}} \\

& RAG
& 0.5392(\textbf{↑}) & 0.3082(\textbf{↑})
& 0.4521(\textbf{↑}) & 0.5548(\textbf{↑}) & 0.3425(\textbf{↑})
& 0.2397(\textbf{↑}) & 0.0616(\textbf{↑})
& 0.0342(\textbf{↑}) & 0.9259(\textbf{↓}) \\

& PAG
& 0.5537(\textbf{↑}) & 0.2705(\textbf{↑})
& 0.5959(\textbf{↑}) & 0.4863(\textbf{↑}) & 0.3082(\textbf{↑})
& 0.1918(\textbf{↑}) & 0.0548(\textbf{↑})
& 0.0411(\textbf{↑}) & 0.8734(\textbf{↓}) \\

\cmidrule(lr){2-11}

& \multicolumn{10}{c}{\textit{\textbf{Memory-based Methods}}} \\

& MemoryBank
& 0.5991(\textbf{↑}) & 0.2568(\textbf{↑})
& 0.7192(\textbf{↑}) & 0.4863(\textbf{↑}) & 0.3904(\textbf{↑})
& 0.1438(\textbf{↑}) & 0.0274(\textbf{↓})
& 0.0137(\textbf{↓}) & 0.9243(\textbf{↓}) \\

& PersonaAgent
& 0.5803(\textbf{↑}) & 0.2842(\textbf{↑})
& 0.7671(\textbf{↑}) & 0.5274(\textbf{↑}) & 0.4658(\textbf{↑})
& 0.1370(\textbf{↑}) & 0.0411(\textbf{↓})
& 0.0137(\textbf{↓}) & 0.8368(\textbf{↓}) \\

& Agentic Memory
& 0.5064(\textbf{↑}) & 0.2603(\textbf{↑})
& 0.5137(\textbf{↑}) & 0.4658(\textbf{↑}) & 0.2877(\textbf{↑})
& 0.1370(\textbf{↑}) & 0.0548(\textbf{↑})
& 0.0342(\textbf{↑}) & 0.8687(\textbf{↓}) \\

\bottomrule
\end{tabular}%
}
\caption{
Average accuracy on ``easy'' and ``hard'' tasks for
gemini-3.1-flash-lite (Gemini) and deepSeek-v4-flash (Deepseek).
"\textbf{↓}" and "\textbf{↑}" indicate decreases and improvements relative to the baseline respectively.
}
\label{tab:improvement_methods}
\end{table*}
\section{Experiments}

\subsection{Evaluation Settings}

\paragraph{Baseline Models}
Fifteen mainstream LLMs from diverse model families and capability tiers are evaluated. Including OpenAI GPT series\cite{gpt5},Anthropic Claude series\cite{claudesonnet5}, Google Gemini series\cite{gemini3.1,gemini3.5}, xAI Grok series\cite{grok4}, DeepSeek-AI DeepSeek series\cite{deepseekv4,deepseekv3.2,deepseekv3}, Alibaba Qwen series\cite{qwen3} and Xiaomi mimo series\cite{mimov2.5}. For models with multiple thinking intensities, LLMs with different modes of thinking are set to the "default" thinking intensity to ensure fairness and reproducibility. We run experiments on a severwith 48-core Intel CPU and 4 NVIDIA RTX 3090 GPUs. LLMs are evaluated by API calling and the temperature is set to 0 for reproducibility, check appendix for details.

\paragraph{Experimental Methods}
The effects of personalization and memory augmentation are examined on gemini-3.1-flash-lite and deepseek-v4-flash using seven settings organized into four categories. In the unaugmented \emph{Baseline}, the complete implicit interaction history is provided without explicit preference reminders, retrieval, or memory transformation. The prompt-based \emph{Reminder} method additionally inserts a fixed instruction to recall and follow user preferences before plan definition, plan execution, and plan update. Among retrieval-based methods, \emph{RAG}\cite{rag} retrieves the top-3 task-relevant historical interactions, while \emph{PAG}\cite{pag} supplements the same evidence with a static, task-conditioned user profile derived from the complete history. Among memory-based methods, \emph{MemoryBank}\cite{memorybank} adds independently summarized interaction memories to the retrieval pool; \emph{PersonaAgent}\cite{personaagent} augments PAG with task-relevant external knowledge to support action selection; and \emph{Agentic Memory}\cite{vitabench2} maintains a versioned user profile that is conservatively updated by the evaluated model after each planning, action, and update turn. All saved trajectories are assessed using the same fine-grained binary rubrics. Check appendix for more information about the methods.

\subsection{Main Result}

\paragraph{Current LLMs can instantiate preferences locally, but cannot reliably maintain an executable and preference-consistent dynamic plan.} As shown in Table~ \ref{tab:plan_performance}, across the 15 models, the average preference score is 49.27\%, whereas the average quality score is 29.08\%. Even gpt-5.6-terra, which leads both metrics, achieves only 66.32\% in preference score. Overall, existing models remain substantially limited in their ability to define and update plans.

\paragraph{Baseline Performance Analysis}
Plan definition remains challenging and plan update constitutes the primary bottleneck, and plan execution is comparatively strong. At plan definition, average preference and quality accuracies reach 51.51\% and 48.07\%, but only 31.86\% joint accuracy is achieved, indicating that preferences, objectives, dependencies, and feasibility constraints are rarely integrated simultaneously. After tool feedback, average preference, quality, and joint accuracies fall to 18.80\%, 10.08\%, and 7.02\%, respectively, as invalidated steps are seldom revised without disrupting valid constraints. By contrast, execution preference accuracy averages 77.52\%. Thus, reliable preference grounding in execution does not imply lifecycle-level planning competence.

\paragraph{Impact of Task Difficulty}
Similarly low personalized tool-use planning performance is observed for mainstream LLMs across task difficulty levels. For plan execution, trajectory correctness (Traj.) is manually measured to assess execution stability. As shown in Figure~\ref{fig:difficulty}, deepseek-v4-pro and mimo-v2.5-pro exhibit closely aligned performance on the easy and hard subsets across all three planning stages and the Traj. metric, with no consistent advantage on easier tasks. This consistency indicates that the observed deficiencies are systematic limitations of current LLMs and are largely independent of task difficulty.

\subsection{Analysis of Improved Methods}
\textbf{Improved methods can strengthen plan definition and Plan Update, but substantially weaken preference adherence during Plan Execution.} As shown in Table~\ref{tab:improvement_methods}, Definition joint accuracy is improved by every method, and Update gains are obtained in several settings. However, execution preference accuracy is reduced across both models under all tested methods. The baseline imbalance is therefore reversed rather than resolved: stronger plan formulation and revision are accompanied by less reliable preference grounding in tool arguments. Existing methods thus fail to preserve their planning gains throughout the complete lifecycle.

\subsection{Error Analysis}
\begin{table}
\centering
\setlength{\tabcolsep}{3.2pt}
\renewcommand{\arraystretch}{1.08}

\resizebox{\columnwidth}{!}{%
\begin{tabular}{@{}c|c|ccc@{}}
\toprule
\multirow{2}{*}{\textbf{Model}}
& \multirow{2}{*}{\textbf{Error Type}}
& \multicolumn{3}{|c}{\textbf{Planning Stage}} \\
\cmidrule(lr){3-5}
& & \textbf{Update} & \textbf{Definition} & \textbf{Execution} \\
\midrule

\multirow{4}{*}{\shortstack{Deepseek\\v4-pro}}
& \preferrorcell{E1: Preference omission}
& \preferrorcell{79.9}
& \preferrorcell{43.9}
& \preferrorcell{23.4} \\

& \preferrorcell{E2: Preference conflict}
& \preferrorcell{89.3}
& \preferrorcell{10.7}
& \preferrorcell{38.3} \\

& \taskerrorcell{E3: Goal forgetting}
& \taskerrorcell{86.0}
& \taskerrorcell{7.5}
& \taskerrorcell{--} \\

& \taskerrorcell{E4: Logic error}
& \taskerrorcell{93.5}
& \taskerrorcell{47.2}
& \taskerrorcell{--} \\

\midrule

\multirow{4}{*}{\shortstack{Gemini-3.1\\flash-lite}}
& \preferrorcell{E1: Preference omission}
& \preferrorcell{81.3}
& \preferrorcell{75.0}
& \preferrorcell{22.2} \\

& \preferrorcell{E2: Preference conflict}
& \preferrorcell{91.0}
& \preferrorcell{18.1}
& \preferrorcell{32.6} \\

& \taskerrorcell{E3: Goal forgetting}
& \taskerrorcell{88.2}
& \taskerrorcell{53.5}
& \taskerrorcell{--} \\

& \taskerrorcell{E4: Logic error}
& \taskerrorcell{92.4}
& \taskerrorcell{78.5}
& \taskerrorcell{--} \\

\bottomrule
\end{tabular}%
}

\caption{Error-type prevalence among failed instances (\%) for the two baseline LLMs.``--'' denotes not applicable.}
\label{tab:baseline_core_errors}
\end{table}

To identify the main drivers of agent failure in PDEU-Bench, observed errors are grouped into four non-mutually-exclusive categories: (E1) \textit{Preference Omission}, denoting an omitted or unretained preference; (E2) \textit{Preference Conflict}, denoting a direct violation of an explicit or inferred preference; (E3) \textit{Goal Forgetting}, denoting the loss of task objectives or success criteria; and (E4) \textit{Logic Error}, denoting infeasible steps, invalid dependencies, or inconsistent revisions. Multiple categories may be assigned to a single failed trajectory. Since Plan Execution assesses preference grounding in tool arguments rather than plan structure, only E1 and E2 are applicable at this stage. 

\paragraph{Baseline Error Analysis}
Both plan definition and plan update exhibit substantial preference-following failures, manifested as preference omission and preference conflict. As shown in Table~\ref{tab:baseline_core_errors}, during plan definition, preference omission affects 43.9--75.0\% of failed instances, while preference conflicts occur in 10.7--18.1\%. Plan-quality errors are also evident at this stage: goal forgetting affects 7.5--53.5\% of failed instances, and logic errors occur in 47.2--78.5\%. These failures become substantially more severe during plan update, where preference omission and preference conflict reach 79.9--81.3\% and 89.3--91.0\%, respectively; meanwhile, goal forgetting rises to 86.0--88.2\% and logic errors to 92.4--93.5\%. Even during the comparatively reliable plan execution stage, 22.2--38.3\% of failed instances remain affected by preference-related errors. 

\begin{figure}
    \centering
    \includegraphics[width=1\linewidth]{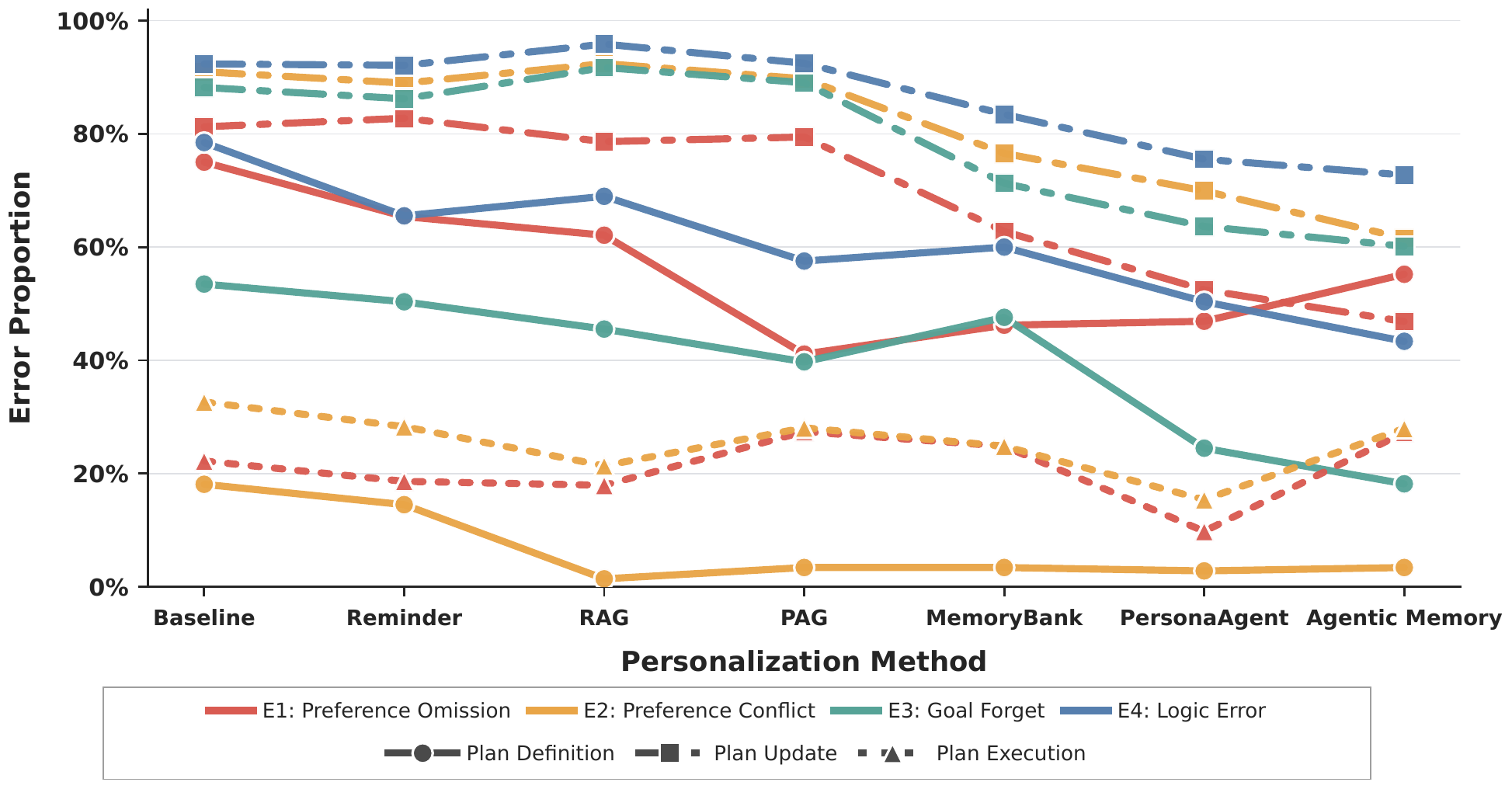}
    \caption{Error Proportion of 6 improved methods, tested on gemini-3.1-flash-lite. The circular, square, and triangular dots represent the error distribution in plan definition, plan update, and plan execution, respectively.}
    \label{fig:method_error}
\end{figure}

\paragraph{Improved Methods Error Analysis}
The ability of mainstream methods to mitigate personalized preference-following errors remains limited. As shown in Figure~\ref{fig:method_error}, more consistent reductions are achieved by memory-based methods than by prompting- or retrieval-based methods, particularly during plan definition and plan update; however, these gains are not reliably transferred to Plan Execution. Consequently, none of the evaluated methods consistently preserves user preferences throughout the complete planning lifecycle.

\section{Conclusion}

We introduce PDEU-Bench, a multi-domain benchmark for evaluating personalized tool-using agents across Plan Definition, tool-grounded Plan Execution, and feedback-driven Plan Update. Its 214 long-horizon tasks span 12 everyday domains and 94 tools, with an explicit, versioned plan treated as the central evaluation object. Experiments on 15 mainstream LLMs reveal a pronounced gap between local preference grounding and lifecycle planning: initial plans remain brittle, and Plan Update constitutes the principal bottleneck. Six personalization and memory-augmentation methods yield stage-specific gains that do not reliably transfer across the planning lifecycle. Error analysis further identifies persistent preference omission, preference conflict, goal forgetting, and logical inconsistency. PDEU-Bench provides a reproducible testbed for developing agents that incorporate environmental feedback while maintaining user preferences, task objectives, and plan dependencies over long-horizon interactions.

\bibliography{reference}

\end{document}